\documentclass[letterpaper, 10 pt, conference]{ieeeconf}  

\IEEEoverridecommandlockouts                              

\title{\LARGE \bf Sample Efficient Dynamics Learning for Symmetrical Legged Robots: Leveraging Physics Invariance and Geometric Symmetries }

\author{Jee-eun Lee$^{1}$ and Jaemin Lee$^{3}$ and Tirthankar Bandyopadhyay$^{2}$ and Luis Sentis$^{1}$
\thanks{$^{1}$Jee-eun Lee is with Human Centered Robotics Laboratory, Department of Aerospace Engineering and Engineering Mechanics, The University of Texas at Austin, USA,
        {\tt\small jelee@utexas.edu}}%
\thanks{$^{3}$Jaemin Lee is with the Department of Mechanical and Civil Engineering, California Institute of Technology, Pasadena, CA, USA,
        {\tt\small jaemin87@caltech.edu}}%
\thanks{$^{1}$Luis Sentis is with Faculty of Department of Aerospace Engineering and Engineering Mechanics, The University of Texas at Austin, USA,
        {\tt\small lsentis@austin.utexas.edu}}%
\thanks{$^{2}$Tirthankar Bandyopadhyay is with Robotics and Autonomous Systems Group, Data61, CSIRO, QLD 4069, Australia
        {\tt\small tirtha.bandy@csiro.au}}%
}

\usepackage{times}
\usepackage{multirow}
\usepackage{tabularx}
\usepackage{scrextend}
\usepackage{caption}
\usepackage{subfig}
\usepackage{wrapfig}
\usepackage{lipsum}
\usepackage{amsmath}
\usepackage{amssymb}
\usepackage{algorithm2e}
\usepackage{graphicx}
\usepackage{flushend}

\makeatletter
\renewcommand\p@subfigure{\theFig.~}
\makeatother
\usepackage{color}

\newcommand{\node}{\textbf{n}}
\newcommand{\edge}{\textbf{e}}
\newcommand{\gb}{\textbf{u}}

\newcommand{\bq}{\mathbf{q}}
\newcommand{\bqdot}{\dot{\mathbf{q}}}
\newcommand{\bqddot}{\ddot{\mathbf{q}}}

\newcommand{\bg}{\mathbf{g}}
\newcommand{\bp}{\mathbf{p}}
\newcommand{\ba}{\mathbf{a}}
\newcommand{\bx}{\mathbf{x}}

\newcommand{\btau}{\boldsymbol{\tau}}
\newcommand{\bomega}{\boldsymbol{\omega}}

\begin{document}

\maketitle

\begin{abstract}%
Model generalization of the underlying dynamics is critical for achieving data efficiency when learning for robot control. This paper proposes a novel approach for learning dynamics leveraging the symmetry in the underlying robotic system, which allows for robust extrapolation from fewer samples. Existing frameworks that represent all data in vector space fail to consider the structured information of the robot, such as leg symmetry, rotational symmetry, and physics invariance. As a result, these schemes require vast amounts of training data to learn the system's redundant elements because they are learned independently. Instead, we propose considering the geometric prior by representing the system in symmetrical object groups and designing neural network architecture to assess invariance and equivariance between the objects.Finally, we demonstrate the effectiveness of our approach by comparing the generalization to unseen data of the proposed model and the existing models. We also implement a controller of a climbing robot based on learned inverse dynamics models. The results show that our method generates accurate control inputs that help the robot reach the desired state while requiring less training data than existing methods.
\end{abstract}

\section{Introduction}
Various types of legged systems have been developed with the ability to traverse extreme terrains to increase the versatility of autonomous robots. 
Model-based approaches have been extensively used for controlling these robots \cite{sentis2005synthesis}, \cite{kim2020dynamic}, \cite{orin2013centroidal}, \cite{dai2014whole}. However, such strategies often fail due to complex and unpredictable effects generated from dynamic contacts and other environmental interactions. Also, model-based control requires good dynamic models for accurate trajectory tracking. Alternatively, model-free approaches have shown remarkable success in this field recently. Using the latest deep reinforcement learning techniques, robots can learn complex tasks \cite{levine2016end}, \cite{gu2017deep}, and locomotion \cite{hwangbo2019learning},  \cite{lee2020learning}, \cite{li2021reinforcement}.

However, the lack of physical plausibility in learned models limits their applicability to the vicinity of the training data. Hence, the efficacy of the model depends heavily on the sample complexity. To mitigate this limitation, exhibiting combinatorial generalization is also suggested as a key aspect for modern AI to resemble human intelligence, which can generalize beyond one's experience \cite{Battaglia2018}. There are several network architectures proposed to incorporate prior knowledge. One example is utilizing a differential equation form to encode a physics prior in the Euler-Lagrange equation \cite{lutter2019deep}, \cite{gupta2020structured}. By constraining the model to satisfy a differential equation, it extrapolates more accurately to unseen samples than Feed Forward Neural Networks (FFNNs). Using a mixture of experts (MOE) composed of a system equation (white-box) and a deep neural network (black-box) to learn complex dynamics model is proposed in \cite{ahn2021nested}. It leverages model plausibility from the white-box model while the black-box eliminates the residual error. 

\begin{figure}[t]
\centering
    \hspace{10mm}
    \includegraphics[width=\linewidth]{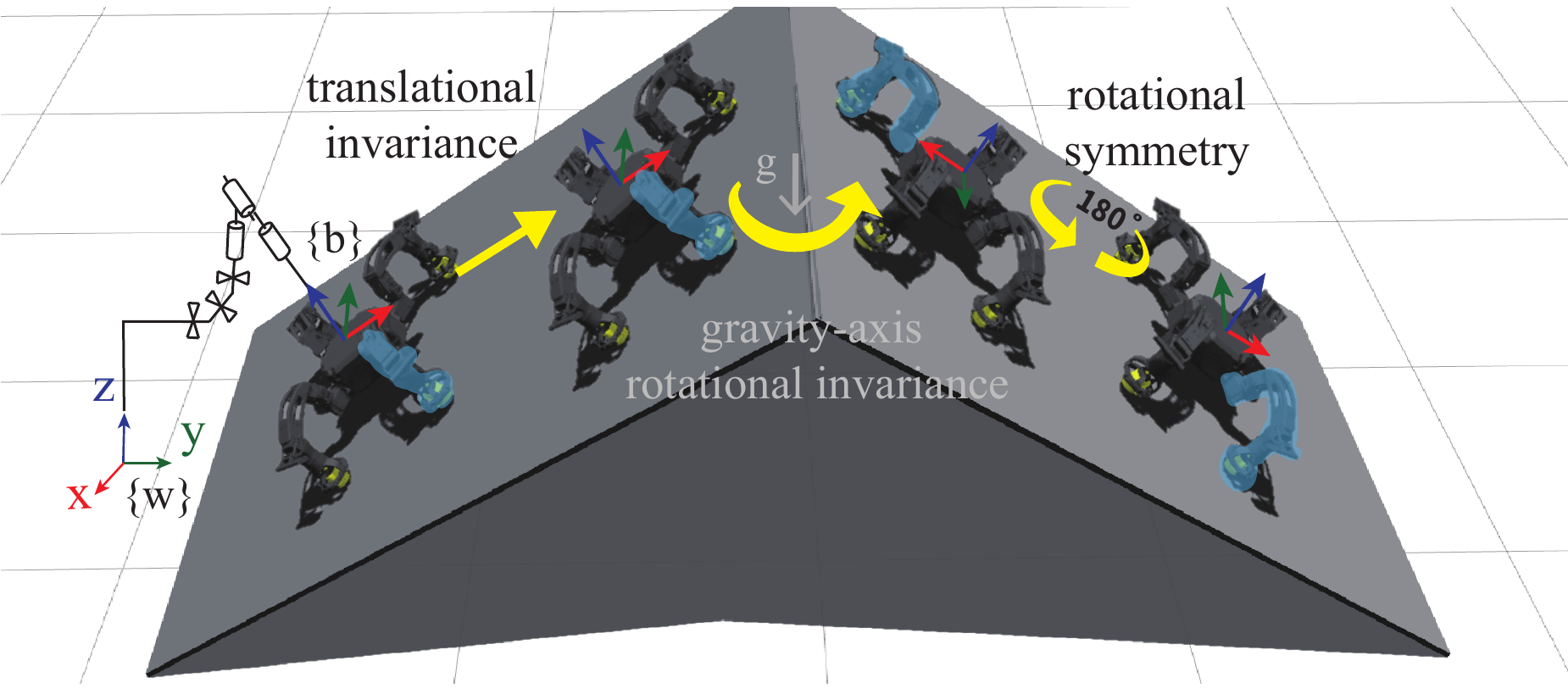}
    \captionof{figure}{The figure shows examples of physical invariance found in a floating-based robot where the legs are symmetrical. Robots in different configurations in the figure above have the same physical properties. e.g., the torque required at each joint to track the given motion is the same. }    \label{fig:intro_invariance}
    \vspace{-6mm}
\end{figure}

Every floating base robot has translation invariance and gravity-axis rotational invariance in dynamics as shown in Fig.~\ref{fig:intro_invariance}.Specifically, multi-legged robots with rotational symmetry like Magneto \cite{bandyopadhyay2018magneto} have a set of configurations that works under the same physics. However, current schemes that represent the states and actions of the robot as a stacked vector hardly capture these properties, meaning that they only learn these symmetric properties through data, resulting in sample inefficiency and a lack of generalization capability. In this paper, we suggest a network topology that works on structured data--a graph or grouped-by-leg data-- reflecting the robot symmetry.

One popular neural network for structured data learning is graph neural networks (GNNs) \cite{scarselli2008graph},\cite{bronstein2017geometric} that is known for its ability to consider the relations between objects by using graph structure for learning. In robotics, there are some previous works that use GNNs to learn dynamics \cite{Sanchez-Gonzalez2018}, \cite{wang2018nervenet}, \cite{whitman2021learning}. However, most of these researches focus on the model transfer capability of GNNs over various types of robots. Reference \cite{Sanchez-Gonzalez2018} emphasizes that the model can predict the dynamics of a new robot(i.e., where its data has not been used for training) by learning the physical relations between links and joints. Notably, GNNs can even be applied to zero-shot transfer learning for modular robots \cite{wang2018nervenet}, \cite{whitman2021learning}. 

However, asking models to learn all types of robot physics is unrealistic. It would require tremendous training data; it is not applicable to robots with different types of sensors and actuators; and sensors produced by different manufacturers can have different performances. Instead, we focus on reducing required sample complexity for system identification using physical priors. In this paper, we introduce new data representations for symmetrical-legged robots and how GNNs and other group-equivariant neural networks can improve sample efficiency based on the permutation equivariance in legged-robot dynamics.

\subsection{Contribution}
This paper proposes sample efficient dynamics learning for symmetrical-legged robots. By using grouped-by-leg structured data, we can define group invariance and equivariance followed by geometric symmetries of a robot based on the data representation. Then, by training a model based on the networks designed to preserve the group equivariance, we can learn the dynamics that incorporate physics priors, providing sample efficiency and generalization capability.

\section{Preliminaries}
Using \textit{a priori} representational and computational assumptions to achieve generalization is actively studied in the deep learning. This can be realized by representing prior knowledge as equivariance or invariance under group actions. Below we introduce the relevant concepts and prior works.

\subsection{Invariance, Equivariance, and Symmetries}
In geometry, we say that an object has symmetry if the object has an invariance under the group action. In group theory, group action, invariance, and equivariance can be defined as follows: 

\textbf{Definition 1 (Group Action)}~ A group G is said to act on a set $X$ if there is a
map $\psi : G\times X \rightarrow X$ called the Group Action such that $\psi(e, x) = x$ for all $x \in X$,  and
$\psi(g,\psi(h, x)) = \psi(gh, x)$ for all $g,h \in G, x \in X$. For compactness we follow the practice
of denoting a group action by a ‘$\circ$’ e.g. $g \circ x$. 

\textbf{Definition 2 (Invariance)} Let $G$ be a group which acts on the sets $X$. We say that
the function $f : X \rightarrow Y$ is $G$-invariant if $f(g \circ x) = f(x)$ for all $x \in X,g \in G$.  

\textbf{Definition 3 (Equivariance)} Let $G$ be a group which acts on the sets $X$ and
$Y$. We say that the function $f : X \rightarrow Y$ is $G$-equivariant if  
$f(g \circ x) = g \circ (f(x))$ for all $x \in X, g \in G$. 

\subsection{Equivariance and Invariance in Neural Networks}
Many works in deep learning leverage equivariance and invariance for constructing the neural network architecture. Data augmentation is a straightforward way to learn group-equivariant representation \cite{qi2020learning}; however, it is computationally inefficient due to the increased data volume to train the model.\\
\textbf{Convolutional Neural Networks~} One well-known example to consider equivariance and invariance is convolutional neural networks (CNNs) \cite{lecun1989backpropagation}, \cite{krizhevsky2012imagenet}, which utilize convolutional layer for preserving object identity over the translation group. \\
\textbf{Group-Equivariant Neural Networks~} A group convolutional layer is proposed to consider other symmetries like rotation and reflection \cite{lipman2005linear},\cite{cohen2018spherical}. For more general structures, deep symmetry networks \cite{gens2014deep} capture a wide variety of invariances defined over arbitrary symmetry groups using kernel-based interpolation and symmetric space pooling. For a set structure like point clouds, simple permutation-equivariant layers \cite{ravanbakhsh2016deep} are shown to be beneficial.\\
\textbf{Graph Neural Networks~} Graph neural networks \cite{scarselli2008graph}, \cite{Zhou2018} is a type of neural networks directly operating on graph structures where a system of objects and relations is represented through nodes and edges. GNNs can capture all permutation invariance, and equivariance \cite{maron2018invariant} in nodes and edges. Graph Networks (GNs), proposed by \cite{Battaglia2018}, provides a general form of graph structure for learning, $G=(\gb, \{\node_i\}, \{\edge_j,s_j,r_j\})$ that includes a graph embedding vector called global features $\gb$, a set of node features $\{\node_i\}_{i=1,..,N_n}$ and a set of edges $\{\edge_j,s_j,r_j\}_{j=1,..,N_e}$, where $\edge_j$ is a vector representing edge features and $s_j$, $r_j$ are the indices of the sender and receiver nodes ranging from $1$ to $N_n$. This data structure allows GNNs to capture the permutation equivariance of nodes and edges. Then graph networks are defined as a "graph2graph" module, which takes an input graph and returns an output graph with the updated features. 

\section{Structured Representation for Symmetrical Legged Robot Dynamics}
This section describes the equivariance and invariance in inverse dynamics of symmetrical-legged robots. Then, we suggest data representation that can capture those properties in neural networks.

\subsection{Dynamics of Floating-base Legged Robots}
The dynamics of legged robots are commonly described as a function of the generalized coordinates $\bq=[\bq_b^\top, \bq_j^\top]^\top\in \mathbb{R}^{n_q}$, where $\bq_b\in\mathbb{R}^6, \bq_j \in \mathbb{R}^{n_j}$ represent the configuration of the floating base and joints, respectively. 
As such, the rigid-body dynamics of a floating base robot can be obtained using Euler-Lagrange formalism yielding:
\begin{equation}
    \mathbf{M}(\bq) \bqddot + \mathbf{b}(\bq,\bqdot) = \mathbf{S}_{a}^{\top} \btau_a + \mathbf{J}_{c}(\bq)^{\top}\mathcal{F}_{c}
\label{eqn:fbs_dyn}
\end{equation}
where $\mathbf{M}(\bq)$, $\mathbf{b}(\bq,\bqdot)$ represent the mass/inertia matrix and the Coriolis/centrifugal force plus the gravitational force. $\mathbf{S}_{a} \in \mathbb{R}^{n_a \times n_q}$ denotes the selection matrix corresponding to the index set of actuated joints, which maps $\btau_a$ (the actuated joint torques) into the generalized forces and, $\mathbf{J}_c(\bq)$ denotes the stacked contact Jacobian matrix that maps contact wrench vector $\mathcal{F}_{c}$ into the generalized forces. Finally, from the dynamics equation, we can derive the forward model $f_{\textrm{fwd}}$ and inverse model $f_{\textrm{inv}}$ as:
\begin{equation}
\bqddot = f_{\textrm{fwd}}(\bq, \bqdot, \btau), \quad
\btau = f_{\textrm{inv}}(\bq, \bqdot, \bqddot) 
\label{eqn:fwd&inv}
\end{equation}

\subsection{Physics Invariance in Floating-base Robot}
\label{subsec:invariance}
Let $G_\theta$ be the gravity-axis rotation group and $T_p$ be the translation group acting on the base configuration space $\bq_b=[x,y,z,rz,ry,rx]^\top$ as follows:
\begin{align*}
    G_\theta \circ \bq_b &= [x,y,z,rz',ry',rx']^\top, \\
    T_p \circ \bq_b &= [x+p_x,y+p_y,z+p_z,rz,ry,rx]^\top
\end{align*}
where, $rz',ry',rx'$ represent the EulerZYX angle of base configuration that is rotated by $\theta$ along the gravity axis. Similarly, we can represent $\bqdot_b, \bqddot_b$ when the rotation around the gravity axis and translation groups are applied to the robot. Then, the gravity-axis rotational invariance and translational invariance can be represented as:
\begin{align}
\btau &= f_{\textrm{inv}}(\bq,\bqdot,\bqddot) =f_{\textrm{inv}}(\bq_b, \bqdot_b, \bqddot_b, \bq_j, \bqdot_j, \bqddot_j) \\
&= f_{\textrm{inv}}(g \circ (\bq_b, \bqdot_b,  \bqddot_b), \bq_j, \bqdot_j, \bqddot_j ), ~\textrm{for all } g\in\{G_\theta, T_p\}  \nonumber
\end{align}

\subsection{Rotational Equivariance in Symmetric Legged Robots}
\label{subsec:equivariance}
\begin{figure}[t]
  \includegraphics[width=0.95\linewidth]{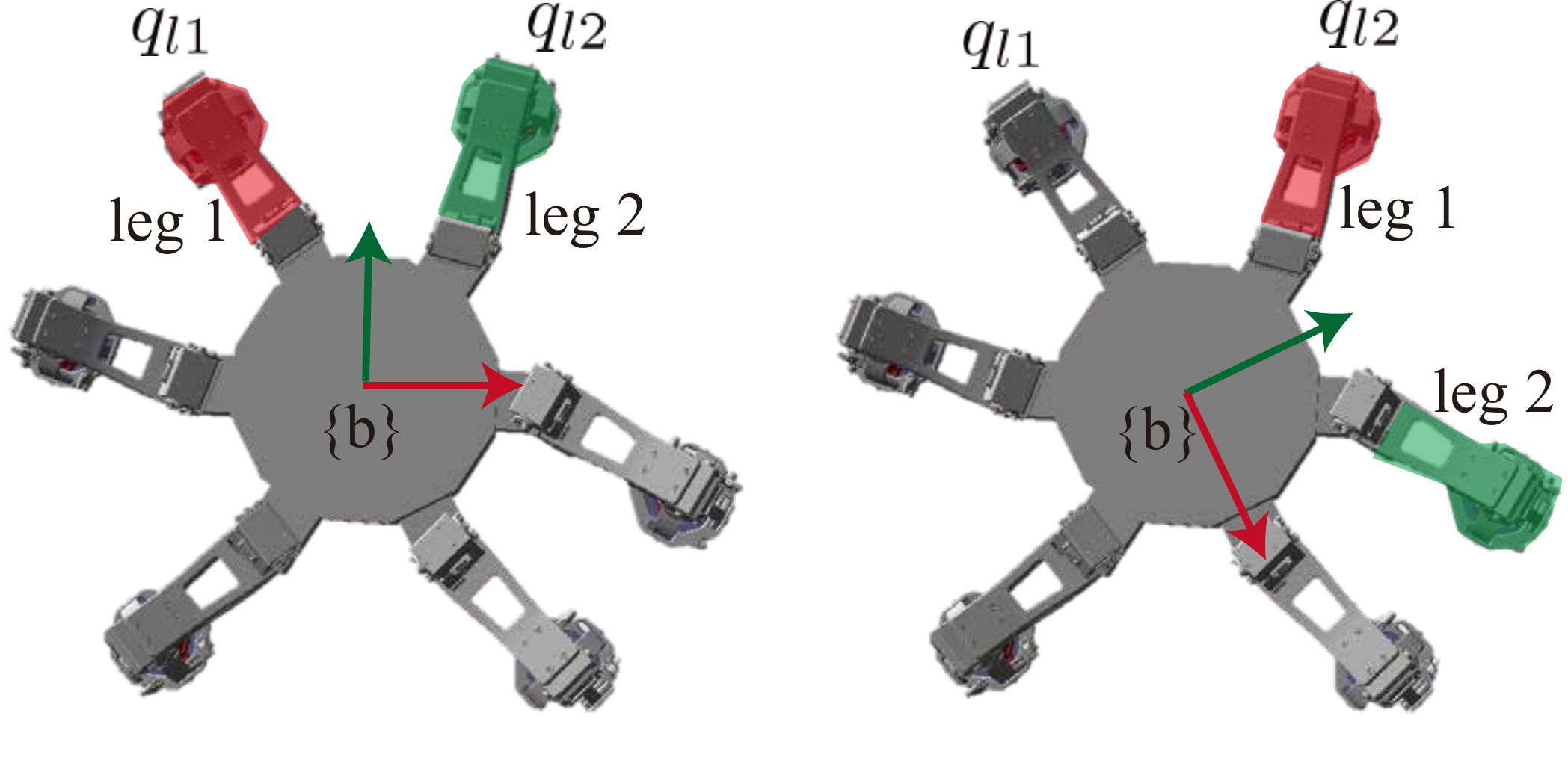}
  \caption{Rotational equivariance in a symmetric hexapod. If we rotate the hexapod by $\frac{\pi}{3}$ and shift the joint values assigned to legs circularly, we can make a set of robot configurations that looks exactly the same. Then the robots have the same physics applied to each leg, seemingly at the same position. }
  \label{fig:rotational_symmetry}
  \vspace{-5mm}
\end{figure}
Let us assume a symmetric legged robot with $n$ legs, which has $n$-th order rotational symmetry. Fig.~\ref{fig:rotational_symmetry} shows an example of $n=6$. Each leg of the robot has $n_{l}$ degrees of freedom and is labeled from $1$ to $n$ in a cyclic order. 
Then, we can split the joint configuration by leg groups: $\bq_j = (\bq_{l1}, \dots , \bq_{ln}) $ where $\bq_{li}\in \mathbb{R}^{n_{l}}$ represents $i$-th leg joint configuration. Similarly, the actuated joint torques can be divided into: $\btau_a = (\btau_{l1},\dots,\btau_{ln})$. Now we define the leg circular shift for the joint configuration and torque as $P_\theta$, which shifts the joint values(e.g. $\bq_{li}, \bqdot_{li}, \bqddot_{li}, \btau_{li}$) assigned to each leg to the neighbouring leg as described in Fig.~\ref{fig:rotational_symmetry}:
\begin{align*}
    P_\theta \circ \bq_j &= P_\theta \circ (\bq_{l1},\bq_{l2}, \dots , \bq_{ln})  =
    (\bq_{l2},\bq_{l3}, \dots , \bq_{l1}) \\
    P_\theta \circ \btau_a &= P_\theta \circ (\btau_{l1},\btau_{l2}, \dots , \btau_{ln}) = 
    (\btau_{l2},\btau_{l3}, \dots , \btau_{l1})
\end{align*}
and by denoting $(P_\theta \circ \bq_j, P_\theta \circ \bqdot_j, P_\theta \circ \bqddot_j )=P_\theta \circ (\bq_j, \bqdot_j, \bqddot_j)$ and the rotation of the base configuration as $R_\theta \circ (\bq_b, \bqdot_b,  \bqddot_b)$, it is clear that inverse dynamics of a robot with circular symmetry satisfies: 
\begin{align}
P_\theta \circ \btau &= P_\theta \circ f_{\textrm{inv}}(\bq_b, \bqdot_b, \bqddot_b, \bq_j, \bqdot_j, \bqddot_j)  \nonumber \\ 
&= f_{\textrm{inv}}(R_\theta \circ (\bq_b, \bqdot_b,  \bqddot_b), P_\theta \circ (\bq_j, \bqdot_j, \bqddot_j) )
\label{eqn:equivariance}
\end{align}

Unfortunately, though it seems there exists a kind of equivariance in symmetric legged robot dynamics, Eqn. (\ref{eqn:equivariance}) shows that the current data representation cannot satisfy the definition of $G$-equivariance exactly, which is $f(g \circ x) = g \circ (f(x))$, due to $\bq_b$ in the input space. 
We introduce a new representation model for the symmetrical-legged robot dynamics system to address this issue in the next section.

\subsection{Robot Representation Model}
\label{sec_3_1}
\begin{figure}[t]
  \includegraphics[width=0.95\linewidth]{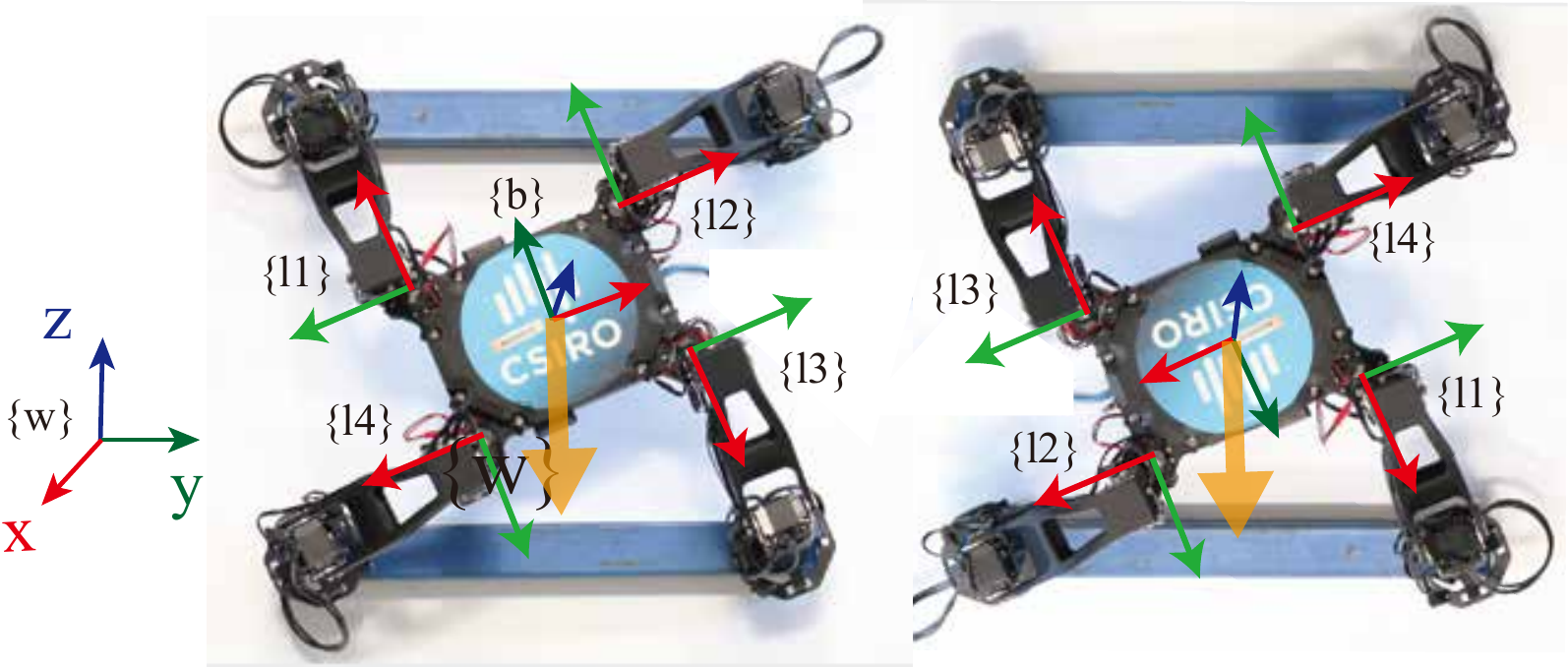}
  \caption{Base link state representation in each leg frame. Given that the leg frames are also in symmetry for the symmetric configuration, the quantities expressed from the leg frames are invariant to the rotation of base link, allowing us to define equivariance over the rotation group.}
  \label{fig:robot_representation}
  \vspace{-5mm}
\end{figure}
The main idea of this representation is first, to configure the domain ($\bq_b, \bq_j$) and codomain ($\btau_a$) of the model as a set of leg configurations ($\{X_{li}\}_{i \in \textrm{leg}}$) and second, the fact that, for the given twist, its body-frame representation does not
depend on the choice of the fixed frame \cite{lynch2017modern}. 
For this purpose, we propose to express the information regarding base configuration from the leg coordinates and put the transformed information into each leg-group. To utilize the invariance and equivariance properties of dynamics addressed in section \ref{subsec:invariance} and \ref{subsec:equivariance} without loss of information, we chose linear acceleration, angular velocity and gravity vector ($\bomega_{b}, \ba_{b}, \bg_{b}$) to represent base configuration information. Then, we have transformed values:
\begin{align}
    \bomega_{i} = R_{ib}\bomega_{b}, \quad \ba_{i} = R_{ib}\ba_{b}, \quad \bg_{i} = R_{ib}\bg_{b}
\end{align}
where $\bomega_i, \bomega_b, \ba_i$ and $\ba_{b}$ represent angular velocities and linear accelerations of the base link expressed in i-th leg frame $\{li\}$ and base frame $\{b\}$, respectively. $\bg_{i}, \bg_{b}$ represent gravity vector expressed from the frame $\{li\}, \{b\}$, and $R_{ib}$ represents rotation matrix that maps a vector from $\{b\}$ to $\{li\}$. We locate the leg frame $\{li\}$ at the first joint of the $i$-th leg at zero configuration as described in Fig.~\ref{fig:robot_representation}, which is attached to the base frame. 

Instead of using ($x,y,z, rz, ry, rx$) to represent base configurations, by adding contact information, we can derive these information from the joint states of each leg for floating-base robots. Also we can benefit from taking values of IMU sensors in real robots for the above choice of data. In addition to ($\bomega_{i}, \ba_{i}, \bg_{i}$), we added $\bp_i$, the position vector from the i-th leg frame to the base link expressed in $\{li\}$ to identify the leg, and contact/adhesion Booleans $b_c, b_m$. Then, we can represent the robot states via leg-group data structure: 
\begin{align}
&\quad (\bq, \bqdot, \bqddot) \triangleq \bx=(\bx_{l1},\bx_{l2},\dots,\bx_{ln}) \label{eqn:leg_states} \\
& \textrm{where}~ \bx_{li} = [\bq_{l}^\top,\bqdot_{l}^\top,\bqddot_{l}^\top, \bomega^\top, \ba^\top, \bg^\top, \bp^\top, b_{c}, b_{m} ]_i^\top \nonumber \\
& \textrm{or}~\bx_{li} = [\bq_{l}^\top,\bqdot_{l}^\top,\bq_{l(d)}^\top,\bqdot_{l(d)}^\top, \bomega^\top, \ba^\top, \bg^\top, \bp^\top, b_{c}, b_{m} ]_i^\top \nonumber
\end{align}
where $\bq_{li}, \bqdot_{li}, \bqddot_{li}$ represent the joint states vectors of the i-th leg. Note that the next step desired states $\bq_{l(d)}, \bqdot_{l(d)}$ can be used instead of $\bqddot_{ji}$ in the discrete time system. Then we can rewrite the inverse dynamics as:
\begin{align}
\btau &= (\btau_{l1},\dots,\btau_{ln}) \nonumber \\
&= f_{\textrm{inv}}(\bq,\bqdot,\bqddot) = f_{\textrm{inv}}(\bx_{l1},\bx_{l2},\dots,\bx_{ln}) = f_{\textrm{inv}}(\bx) \label{eqn:inv_dyn}
\end{align}
Now we show that this data representation satisfies invariance and equivariance properties discussed in section
 \ref{subsec:invariance} and \ref{subsec:equivariance}. 
 
\textbf{Gravity-axis rotational invariance ($G_\theta$)} By representing all the base configuration quantities $(\bg, \bomega, \ba, \bp)_i$ from the frame attached to the base link, the data representation itself is invariant to $G_\theta$. For example, $\bg_{i} = R_{ib}\bg_{b}$ is invariant to rotations about the gravity axis. It's derivation is as follows:
\begin{equation}
\bg_{b} = R_{bw}\bg_w = R_{bw}e^{[\hat{\bg}_w]\theta}\bg_w = R_{b'w}\bg_w = \bg_{b'}
\end{equation}
where $\textbf{g}_w = [0,0,-9.81]^\top$ is the gravity vector expressed in the world frame, and $e^{[\hat{\textbf{g}}_w]\theta} \in SO(3)$ represents the rotation about the axis $\hat{\textbf{g}}_w$ by an angle $\theta$. Suppose that $\{b'\}$ has the same origin as $\{b\}$ but it's rotated by $\theta$ about the gravity vector, we get $R_{b'w}=R_{bw}e^{[\hat{\textbf{g}}_w]\theta}$. Finally, as vectors that are parallel to the rotation axis are not changed by the rotation, we get  $e^{[\hat{\textbf{g}}_w]\theta}\textbf{g}_w = \textbf{g}_w$, and thus $\bg_{i} = R_{ib}\bg_{b}$ is invariant to the gravity-axis rotation group with fixed $R_{ib}$.

\textbf{Translational invariance ($T_\theta$)}  Instead of taking current base link position and velocity as an input, we decided to let the model derive the linear states of the base link from the joint states of legs where its foot is in contact. As a result, our data representation can be invariant to $T_\theta$, .

\textbf{Rotational symmetry equivariance ($P_\theta$)} 
In the proposed representation, if the robots are in symmetry configuration, leg frames locations and orientations are also in symmetry. For example, as shown in Fig. \ref{fig:robot_representation}, 
frame $\{l2\}$ of the left robot and frame $\{l4\}$ of the right are located in the same Configuration. Therefore the base link motion such as linear acceleration, angular velocity and gravity vector expressed from frame $\{l2\}$ of the left and frame $\{l4\}$ of the right will have the same values. Then, we have 
\begin{align*}
 P_\theta \circ \btau &= f_{\textrm{inv}}(P_\theta \circ x) \\
\Leftrightarrow (\btau_{l2},\btau_{l3}, \dots , \btau_{l1})&=f_{\textrm{inv}}(x_{l2},x_{l3},\dots,x_{l1})
\end{align*}
Note that if we define the rotational symmetry equivariance ($P_\theta$) to shift the values assigned to $k$-neighbouring leg instead of the right next leg, we can also describe a robot like Magneto, which has 4 symmetric legs but has 2 order of rotational symmetry, without loss of generality. 


\subsection{Symmetrical Legged Robot Inverse Dynamics Model}
In this paper, we suggest two structured learning to consider symmetries in robot dynamics using GNNs and group-equivariant neural networks. 
\begin{figure}[t]
  \includegraphics[width=\linewidth]{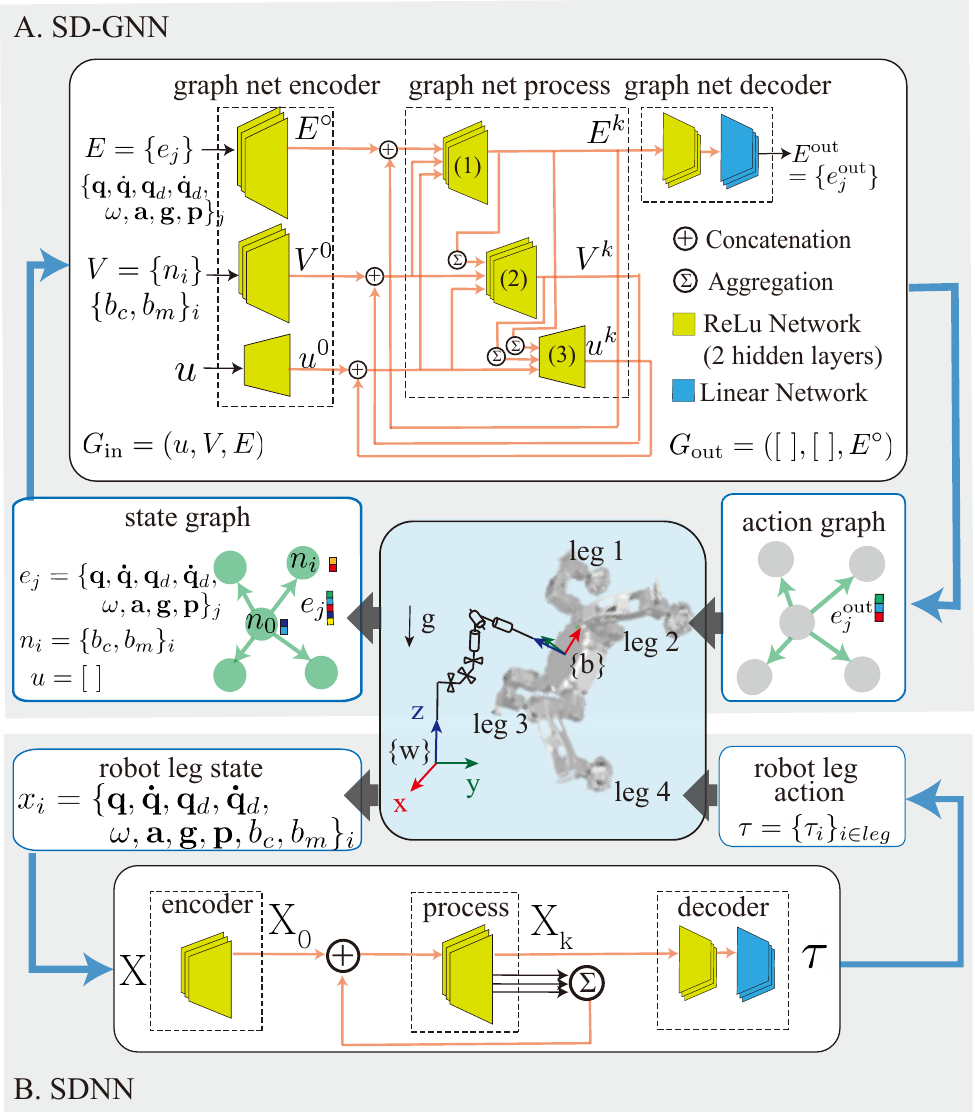}
  \caption{A. Scheme of the inverse dynamics model developed based on SD-GNNs and B. SDNNs working on Magneto.}
  \label{fig:inverse_dynamics_net}
  \vspace{-5mm}
\end{figure}

\textbf{Symmetrical Dynamics GNNs (SD-GNNs)}\\
As explained in Eqn. (\ref{eqn:leg_states}), the domain of the inverse dynamics can be represented in a leg group. In a Graph Networks (GNs), data structure for learning is describes as $G=(\gb, \{\node_i\}, \{\edge_j,s_j,r_j\})$. As shown in Fig.~\ref{fig:inverse_dynamics_net}, by setting the base link and the contact links as nodes and legs as edges, we can represent the domain ($\bx$) and codomain ($\tau$) of the inverse dynamics of legged robots as an input graph ($G_\textrm{in}$) and output graph ($G_\textrm{out}$). We then learn the mapping from states to actions using GNNs. The proposed SD-GNN is constructed based on the \textit{encode-process-decode} architecture introduced in \cite{Battaglia2018}, and overall network scheme is described in Fig.~\ref{fig:inverse_dynamics_net}.

\textbf{Symmetrical Dynamics Neural Networks (SDNNs)}\\
Though GNs provide the general form to represent robots in symmetry, there are unnecessary operations that slow the algorithm and lower the training accuracy. In this paper, we propose symmetrical dynamics neural networks (SDNNs) working on a set structure based on the permutation-equivariant layer similar to \cite{ravanbakhsh2016deep}, \cite{cohen2016group} as described in Fig.~\ref{fig:inverse_dynamics_net}. For the aggregation function, we used a summation operator, which is invariant to the permutation. By leveraging the repeated processing step with concatenating aggregated latent space of all legs, it can predict the desired output while preserving the equivariance to the rotational shift.


\section{Evaluation on a Simulator}
In this section, to demonstrate the extrapolation and applicability of the proposed model, we compare the performance of the proposed SD-GNNs and SDNNs with a 3-hidden-layer FFNNs and typical GNNs described in \cite{Sanchez-Gonzalez2018} that does not consider the geometric symmetries in its data representation. We primarily focus on analyzing its generalization given a validation dataset and the model's reliability when it is used as a locomotion controller.


\subsection{Data Generation}

We have generated a set of trajectories of climbing robots in a DART simulation environment \cite{lee2018dart} using a whole body controller (WBC) \cite{sentis2005synthesis}, \cite{kim2020dynamic} for training and validation of the model. Climbing motions can be created by sequencing step motions where the reference swing foot trajectory is a Hermite spline derived from swing period, swing height and $p_\textrm{goal}$. For this experiment, we set the swing period to 0.5 seconds and sample the swing heights at around 0.05m, and goal positions at around 0.1m away from the initial foot position in the forward, backward, left, and right directions. The magnetic adhesion is deactivated and reactivated upon the initiation and completion of a swing phase.

\subsection{Training Procedure}
Each network in the model was implemented on multi-layer perceptrons (MLPs) and ReLu activation. We used the Adam optimizer with $10^{-3}$ to $10^{-4}$ learning rates and a mini-batch size of 64. We trained each model until $0.7 \times$validation error did not exceed the training error more than five times in a row. The latent sizes and the number of hidden layers of each network and the number of processing steps for core networks used for the experiment are as follows:

\begin{table}[h!]
\vspace{-2mm}
\centering
\begin{tabular}{|l|c|c|c|}
\hline
& latent size & 
\begin{tabular}[c]{@{}c@{}} hidden \\ layers num\end{tabular} &   
{\begin{tabular}[c]{@{}c@{}} processing \\ steps num\end{tabular} } \\ \hline
FFNNs        & 256 units    & 3 & -\\ \hline
Typical GNNs & \begin{tabular}[c]{@{}l@{}}global: 16 units\\ edge: 64 units\\ node: 16 units\end{tabular} & 2 & 5 \\ \hline
SD-GNNs      & \begin{tabular}[c]{@{}l@{}}global: 16 units\\ edge: 64 units\\ node: 16 units\end{tabular} & 2 & 2 \\ \hline
SDNNs        & leg: 128 units &  2 & 2 \\ \hline
\end{tabular}
\vspace{-2mm}
\end{table}

\subsection{Generalization Error to Unseen States}

\begin{figure}[h!]
\centering
\includegraphics[width=\linewidth]{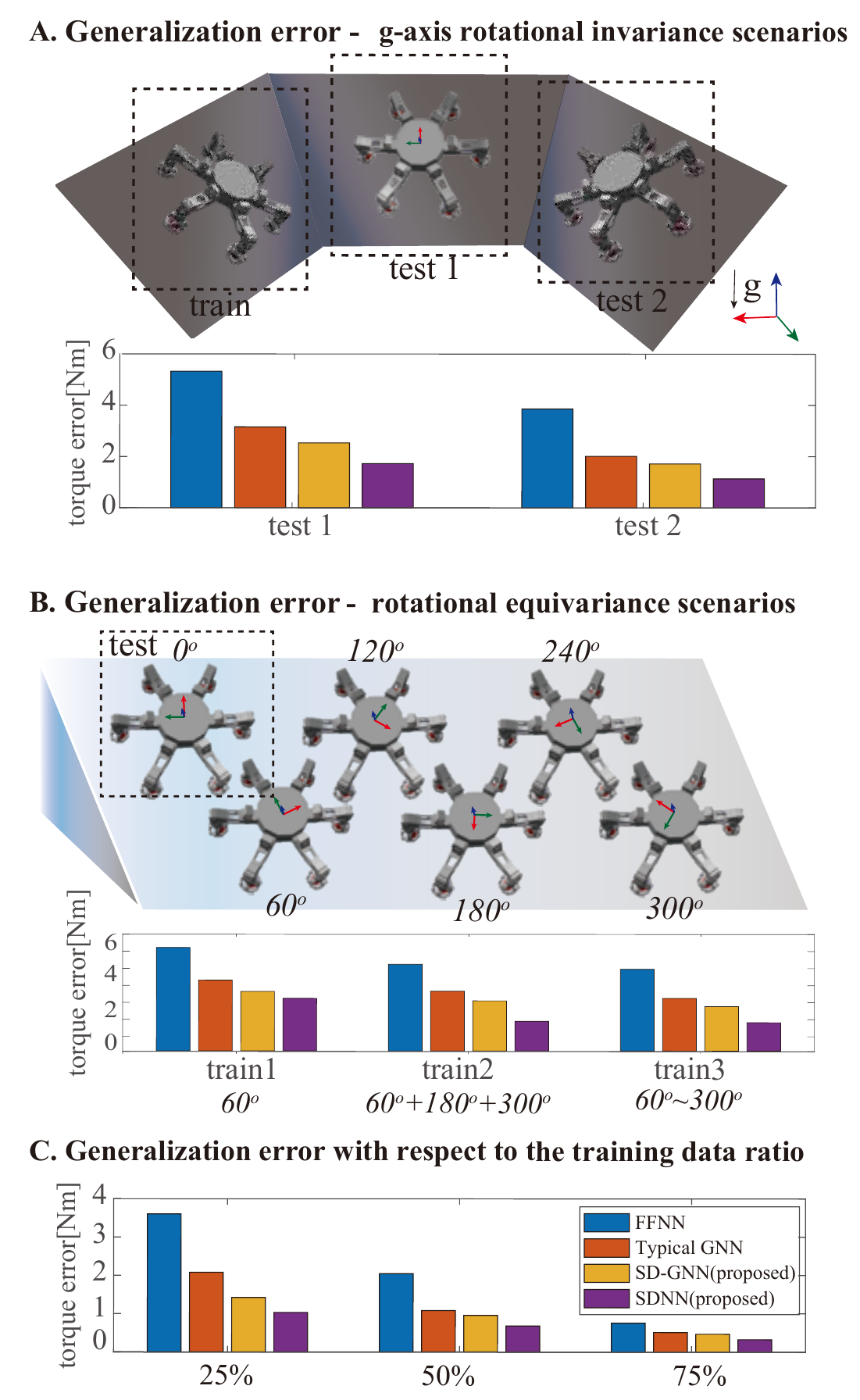}
\caption{SubFig.~A depicts robot climbing scenarios in the various climbing slopes with the same inclination. SubFig.~B shows the generalization ability of each network architecture to predict unseen data in rotational symmetry. SubFig.~C shows sample efficiency of each model.}
\label{fig:generalization_error}
\vspace{-6mm}
\end{figure}

To demonstrate that the proposed SD-GNNs and SDNNs achieve better prediction to unseen data than other networks topology, we run the following experiments for hexa-magneto, which has 6 symmetrical legs. First, we uniformly sample input state and target action tuples to form training and validation datasets from climbing trajectories generated with different slopes and different initial base orientation as shown in Fig.~\ref{fig:generalization_error}. We then train FFNNs, typical GNNs \cite{Sanchez-Gonzalez2018}, SD-GNNs, SDNNs to learn the inverse dynamics model base on the same training dataset and measure the model output error(torque) over the same test data set, which has not been used for training. Fig.~\ref{fig:generalization_error} A and B show the generalization errors over the scenarios of gravitational invariance and rotational equivariance respectively. We can see the proposed SD-GNNs, SDNNs method greatly outperform FFNNs or typical GNNs. Fig.~\ref{fig:generalization_error} C illustrates the sample efficiency and generalization ability of each network by showing the generalization errors with respect to the sample ratio used for training over the whole dataset. We created several training sets by picking scenarios randomly from the whole dataset to train each model and measured the error of the model over the whole data. Fig. \ref{fig:generalization_error}. C. shows that the proposed methods have lower prediction error with less training data, meaning that they yield better generalization ability and are sample efficient.


\subsection{Generalization Performance on Various Robots}
\begin{figure}[t]
    \vspace{1mm}
     \centering
     \includegraphics[width=\linewidth]{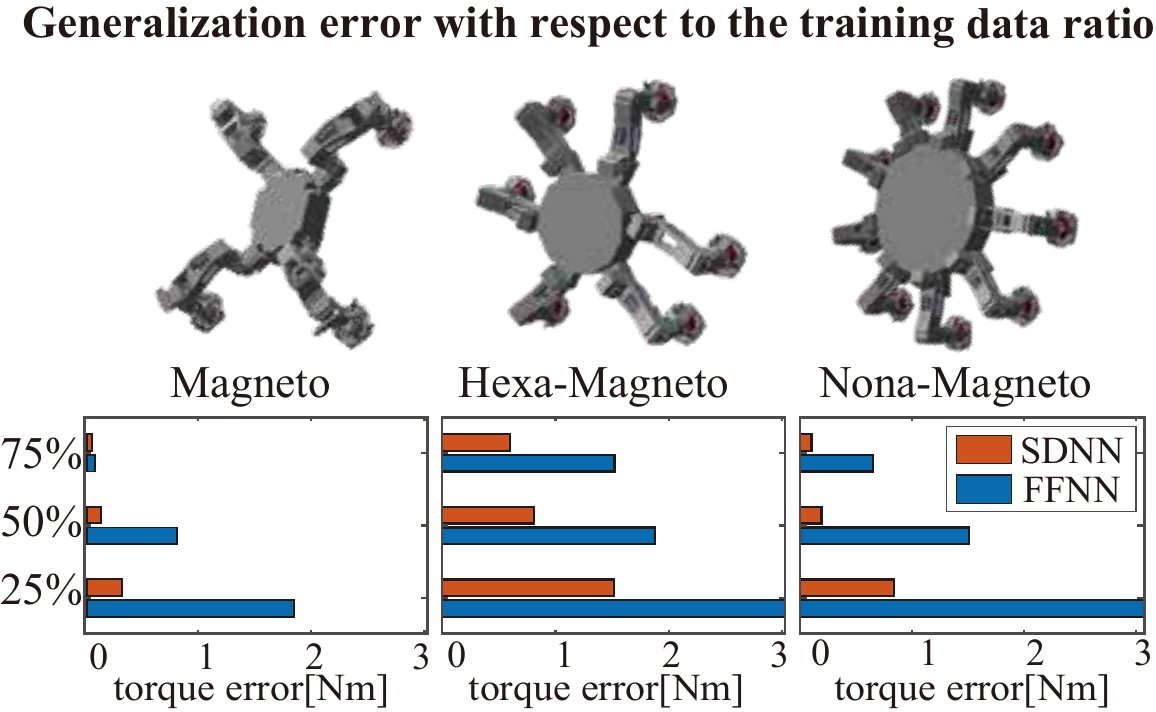}
     \caption{Generalization capability comparison between FFNN and SDNN on robots with different number of legs.  }
     \label{fig:legnum_comapre}
     \vspace{-5mm}
\end{figure}
We investigate the generalization capability of FFNN and the proposed SDNN for three different legged robots, Magneto, Hexa-Magneto, and Nona-Magneto, which have 4, 6, and 9 legs, respectively. As shown in Fig.~\ref{fig:legnum_comapre}, SDNN consistently outperforms FFNN in predicting unseen data with fewer training data. One interesting fact is that in robots with more legs, as the system gets more complicated, FFNN is shown to hardly learn the dynamics model compared to SDNN even with 75\% of the data, while the proposed SDNN works much better. As a result, we can conclude that when the system's degree of freedom increases, SDNN is likely to require less sample complexity than FFNN.   
\subsection{Inverse Dynamics Control Performance}

\begin{figure}[t]
     \centering
     \vspace{1mm}
     \includegraphics[width=\linewidth]{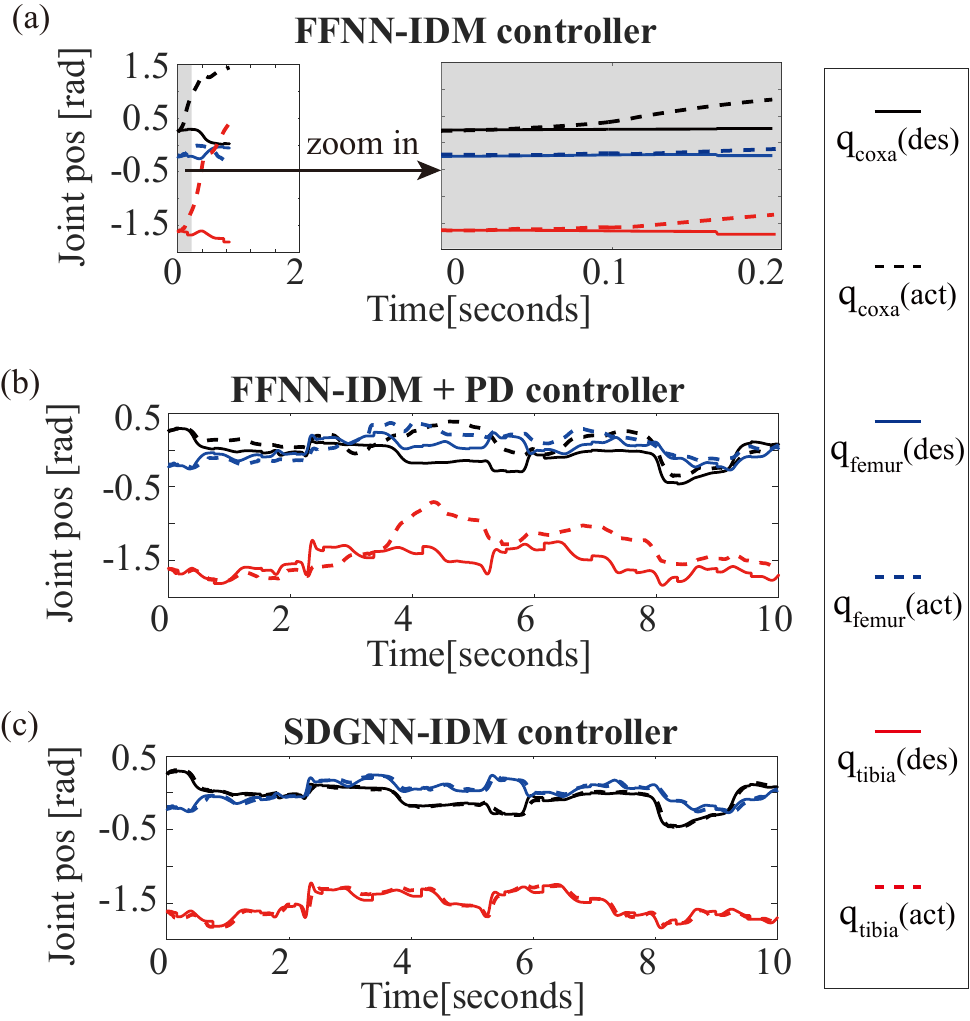}
     \caption{Active joint positions (coxa, femur, tibia joints) of one of Magneto’s legs in a control simulation are shown. Various controllers are designed to track reference climbing trajectories}
     \label{fig:control}
     \vspace{-5mm}
\end{figure}

Aiming to see the performance when each neural network is used as an inverse dynamics controller to track the reference trajectory, we implement two controllers based on SDGNN-IDM (symmetrical dynamics graph neural networks - inverse dynamics model) and FFNN-IDM (feedforward neural networks - inverse dynamics model) working on Magneto. Since we use as inputs to the model the desired states, $\bq_d$ and $\dot \bq_d$ \--- see Fig.~\ref{fig:inverse_dynamics_net} \--- we can use the model directly as a trajectory tracking controller, where the control inputs, $\btau$, obtained from the model is readily applied to the robot. At each time step, the current and desired states of the robot are passed to the model. Then, we control the robot using the outputs of each model. Fig.~\ref{fig:control} shows the tracking performance for each controller. Even though the training loss of both models were similarly small, there were huge differences when they were used as trajectory tracking controllers. As shown in Fig.~\ref{fig:control} {(a)} and \ref{fig:control} {(c)}, the FFNN-IDM based controller fails to track the trajectory within 50ms, while the SDGNN-IDM based controller tracks the trajectory very well even after 10s. To determine why the FFNN-IDM based controller fails, we also implemented a PD augmented FFNN-IDM based controller (Fig.~\ref{fig:control} {(b)}). We set PD gains small enough not to dominate the tracking accuracy. We conclude that the FFNN-IDM based controller failed due to its lack of robustness. On the other hand, SDGNN-IDM works surprisingly well without any additional help of a  traditional feedback controller (Fig.~\ref{fig:control} {(c)}). 

\begin{figure}[t]
    \centering
    \includegraphics[width=\linewidth]{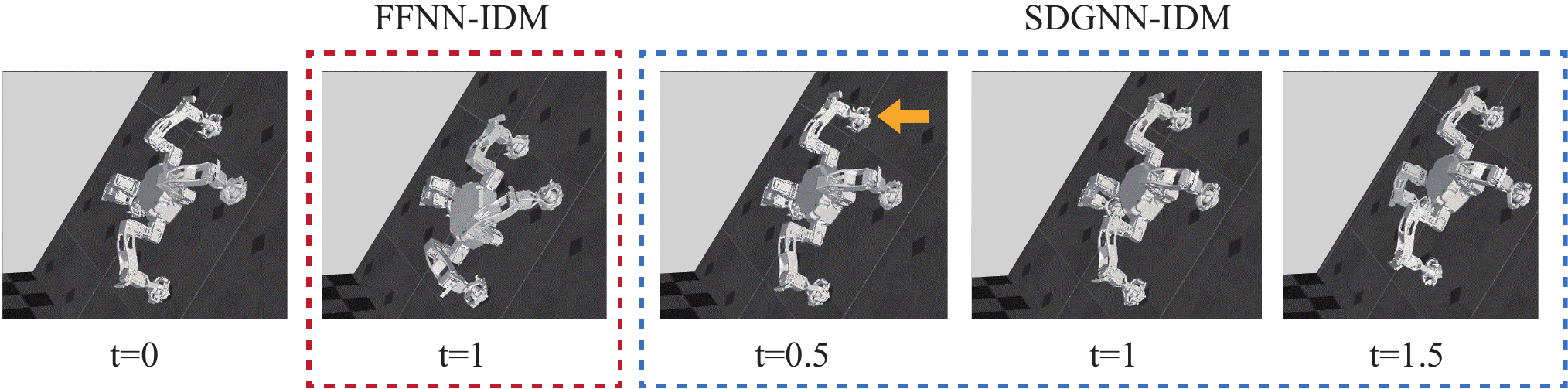}
    \caption{Snapshots of Magneto climbing in simulation. Yellow arrows in the figures indicate the swing foot.}
    \label{fig:snapshots}
    \vspace{-5mm}
\end{figure}


\section{Discussion and Conclusion}
In this paper, we introduce sample efficient robot dynamics learning exploiting the symmetries and physical invariance of the underlying robot. 
Structured data representation allows us to map the identical dynamics resulting from symmetric legs.
Group-equivariant neural network architectures provide the means to incorporate identical dynamical structures in multi-legged robots. As such, it brings substantial efficiency benefits for dynamics learning and generalization capability. 
We have empirically validated the above claims from the following perspectives (i) the proposed SD-GNNs and SDNNs can make more accurate predictions on unseen data, and (ii) SD-GNNs-based inverse dynamics controllers are robust and accurate for trajectory tracking.

A future avenue of research will verify the applicability of the proposed symmetrical dynamics learning for robots with different geometric shapes. Though the proposed model only can capture the radial symmetry in robots, bi-symmetrical robots are also typical such as humanoids and quadrupeds in the legged robots community. We can also evaluate the model's robustness to the unseen environment by examining its performance under the noise and unknown environment properties such as friction, external force, and uneven terrain. We are also interested in knowing how the generalization of our proposed method can help learn robust control policy for data intensive tasks in reinforcement learning. 

\section*{ACKNOWLEDGMENT}
This work has been partially supported by the Office of Naval Research, ONR Grant N000141912311, and by DATA61, CSIRO.

\bibliographystyle{IEEEtran}
\bibliography{root}

\begin{thebibliography}{10}
\providecommand{\url}[1]{#1}
\csname url@rmstyle\endcsname
\providecommand{\newblock}{\relax}
\providecommand{\bibinfo}[2]{#2}
\providecommand\BIBentrySTDinterwordspacing{\spaceskip=0pt\relax}
\providecommand\BIBentryALTinterwordstretchfactor{4}
\providecommand\BIBentryALTinterwordspacing{\spaceskip=\fontdimen2\font plus
\BIBentryALTinterwordstretchfactor\fontdimen3\font minus
  \fontdimen4\font\relax}
\providecommand\BIBforeignlanguage[2]{{%
\expandafter\ifx\csname l@#1\endcsname\relax
\typeout{** WARNING: IEEEtran.bst: No hyphenation pattern has been}%
\typeout{** loaded for the language `#1'. Using the pattern for}%
\typeout{** the default language instead.}%
\else
\language=\csname l@#1\endcsname
\fi
#2}}

\bibitem{sentis2005synthesis}
L.~Sentis and O.~Khatib, ``Synthesis of whole-body behaviors through
  hierarchical control of behavioral primitives,'' \emph{International Journal
  of Humanoid Robotics}, vol.~2, no.~04, pp. 505--518, 2005.

\bibitem{kim2020dynamic}
D.~Kim, S.~J. Jorgensen, J.~Lee, J.~Ahn, J.~Luo, and L.~Sentis, ``Dynamic
  locomotion for passive-ankle biped robots and humanoids using whole-body
  locomotion control,'' \emph{The International Journal of Robotics Research},
  vol.~39, no.~8, pp. 936--956, 2020.

\bibitem{orin2013centroidal}
D.~E. Orin, A.~Goswami, and S.-H. Lee, ``Centroidal dynamics of a humanoid
  robot,'' \emph{Autonomous robots}, vol.~35, no.~2, pp. 161--176, 2013.

\bibitem{dai2014whole}
H.~Dai, A.~Valenzuela, and R.~Tedrake, ``Whole-body motion planning with
  centroidal dynamics and full kinematics,'' in \emph{2014 IEEE-RAS
  International Conference on Humanoid Robots}.\hskip 1em plus 0.5em minus
  0.4em\relax IEEE, 2014, pp. 295--302.

\bibitem{levine2016end}
S.~Levine, C.~Finn, T.~Darrell, and P.~Abbeel, ``End-to-end training of deep
  visuomotor policies,'' \emph{The Journal of Machine Learning Research},
  vol.~17, no.~1, pp. 1334--1373, 2016.

\bibitem{gu2017deep}
S.~Gu, E.~Holly, T.~Lillicrap, and S.~Levine, ``Deep reinforcement learning for
  robotic manipulation with asynchronous off-policy updates,'' in \emph{2017
  IEEE international conference on robotics and automation (ICRA)}.\hskip 1em
  plus 0.5em minus 0.4em\relax IEEE, 2017, pp. 3389--3396.

\bibitem{hwangbo2019learning}
J.~Hwangbo, J.~Lee, A.~Dosovitskiy, D.~Bellicoso, V.~Tsounis, V.~Koltun, and
  M.~Hutter, ``Learning agile and dynamic motor skills for legged robots,''
  \emph{Science Robotics}, vol.~4, no.~26, 2019.

\bibitem{lee2020learning}
J.~Lee, J.~Hwangbo, L.~Wellhausen, V.~Koltun, and M.~Hutter, ``Learning
  quadrupedal locomotion over challenging terrain,'' \emph{Science robotics},
  vol.~5, no.~47, 2020.

\bibitem{li2021reinforcement}
Z.~Li, X.~Cheng, X.~B. Peng, P.~Abbeel, S.~Levine, G.~Berseth, and K.~Sreenath,
  ``Reinforcement learning for robust parameterized locomotion control of
  bipedal robots,'' \emph{arXiv preprint arXiv:2103.14295}, 2021.

\bibitem{Battaglia2018}
P.~W. Battaglia, J.~B. Hamrick, V.~Bapst, A.~Sanchez-Gonzalez, V.~Zambaldi,
  M.~Malinowski, A.~Tacchetti, D.~Raposo, A.~Santoro, R.~Faulkner, C.~Gulcehre,
  F.~Song, A.~Ballard, J.~Gilmer, G.~Dahl, A.~Vaswani, K.~Allen, C.~Nash,
  V.~Langston, C.~Dyer, N.~Heess, D.~Wierstra, P.~Kohli, M.~Botvinick,
  O.~Vinyals, Y.~Li, and R.~Pascanu, ``{Relational inductive biases, deep
  learning, and graph networks},'' \emph{arXiv}, pp. 1--40, 2018.

\bibitem{lutter2019deep}
M.~Lutter, C.~Ritter, and J.~Peters, ``Deep lagrangian networks: Using physics
  as model prior for deep learning,'' \emph{arXiv preprint arXiv:1907.04490},
  2019.

\bibitem{gupta2020structured}
J.~K. Gupta, K.~Menda, Z.~Manchester, and M.~Kochenderfer, ``Structured
  mechanical models for robot learning and control,'' in \emph{Learning for
  Dynamics and Control}.\hskip 1em plus 0.5em minus 0.4em\relax PMLR, 2020, pp.
  328--337.

\bibitem{ahn2021nested}
J.~Ahn and L.~Sentis, ``Nested mixture of experts: Cooperative and competitive
  learning of hybrid dynamical system,'' in \emph{Learning for Dynamics and
  Control}.\hskip 1em plus 0.5em minus 0.4em\relax PMLR, 2021, pp. 779--790.

\bibitem{bandyopadhyay2018magneto}
T.~Bandyopadhyay, R.~Steindl, F.~Talbot, N.~Kottege, R.~Dungavell, B.~Wood,
  J.~Barker, K.~Hoehn, and A.~Elfes, ``Magneto: A versatile multi-limbed
  inspection robot,'' in \emph{2018 IEEE/RSJ International Conference on
  Intelligent Robots and Systems (IROS)}.\hskip 1em plus 0.5em minus
  0.4em\relax IEEE, 2018, pp. 2253--2260.

\bibitem{scarselli2008graph}
F.~Scarselli, M.~Gori, A.~C. Tsoi, M.~Hagenbuchner, and G.~Monfardini, ``The
  graph neural network model,'' \emph{IEEE transactions on neural networks},
  vol.~20, no.~1, pp. 61--80, 2008.

\bibitem{bronstein2017geometric}
M.~M. Bronstein, J.~Bruna, Y.~LeCun, A.~Szlam, and P.~Vandergheynst,
  ``Geometric deep learning: going beyond euclidean data,'' \emph{IEEE Signal
  Processing Magazine}, vol.~34, no.~4, pp. 18--42, 2017.

\bibitem{Sanchez-Gonzalez2018}
A.~Sanchez-Gonzalez, N.~Heess, J.~T. Springenberg, J.~Merel, M.~Riedmiller,
  R.~Hadsell, and P.~Battaglia, ``{Graph networks as learnable physics engines
  for inference and control},'' \emph{35th International Conference on Machine
  Learning, ICML 2018}, vol.~10, pp. 7097--7117, 2018.

\bibitem{wang2018nervenet}
T.~Wang, R.~Liao, J.~Ba, and S.~Fidler, ``Nervenet: Learning structured policy
  with graph neural networks,'' in \emph{International conference on learning
  representations}, 2018.

\bibitem{whitman2021learning}
J.~Whitman, M.~Travers, and H.~Choset, ``Learning modular robot control
  policies,'' \emph{arXiv preprint arXiv:2105.10049}, 2021.

\bibitem{qi2020learning}
G.-J. Qi, L.~Zhang, F.~Lin, and X.~Wang, ``Learning generalized transformation
  equivariant representations via autoencoding transformations,'' \emph{IEEE
  Transactions on Pattern Analysis and Machine Intelligence}, 2020.

\bibitem{lecun1989backpropagation}
Y.~LeCun, B.~Boser, J.~S. Denker, D.~Henderson, R.~E. Howard, W.~Hubbard, and
  L.~D. Jackel, ``Backpropagation applied to handwritten zip code
  recognition,'' \emph{Neural computation}, vol.~1, no.~4, pp. 541--551, 1989.

\bibitem{krizhevsky2012imagenet}
A.~Krizhevsky, I.~Sutskever, and G.~E. Hinton, ``Imagenet classification with
  deep convolutional neural networks,'' in \emph{Advances in Neural Information
  Processing Systems}, F.~Pereira, C.~Burges, L.~Bottou, and K.~Weinberger,
  Eds., vol.~25.\hskip 1em plus 0.5em minus 0.4em\relax Curran Associates,
  Inc., 2012.

\bibitem{lipman2005linear}
Y.~Lipman, O.~Sorkine, D.~Levin, and D.~Cohen-Or, ``Linear rotation-invariant
  coordinates for meshes,'' \emph{ACM Transactions on Graphics (TOG)}, vol.~24,
  no.~3, pp. 479--487, 2005.

\bibitem{cohen2018spherical}
T.~S. Cohen, M.~Geiger, J.~K{\"o}hler, and M.~Welling, ``Spherical cnns,''
  \emph{arXiv preprint arXiv:1801.10130}, 2018.

\bibitem{gens2014deep}
R.~Gens and P.~M. Domingos, ``Deep symmetry networks,'' \emph{Advances in
  neural information processing systems}, vol.~27, 2014.

\bibitem{ravanbakhsh2016deep}
S.~Ravanbakhsh, J.~Schneider, and B.~Poczos, ``Deep learning with sets and
  point clouds,'' \emph{arXiv preprint arXiv:1611.04500}, 2016.

\bibitem{Zhou2018}
J.~Zhou, G.~Cui, Z.~Zhang, C.~Yang, Z.~Liu, L.~Wang, C.~Li, and M.~Sun,
  ``{Graph Neural Networks: A Review of Methods and Applications},''
  \emph{arXiv}, pp. 1--22, 2018.

\bibitem{maron2018invariant}
H.~Maron, H.~Ben-Hamu, N.~Shamir, and Y.~Lipman, ``Invariant and equivariant
  graph networks,'' \emph{arXiv preprint arXiv:1812.09902}, 2018.

\bibitem{lynch2017modern}
K.~M. Lynch and F.~C. Park, \emph{Modern robotics}.\hskip 1em plus 0.5em minus
  0.4em\relax Cambridge University Press, 2017.

\bibitem{cohen2016group}
T.~Cohen and M.~Welling, ``Group equivariant convolutional networks,'' in
  \emph{International conference on machine learning}.\hskip 1em plus 0.5em
  minus 0.4em\relax PMLR, 2016, pp. 2990--2999.

\bibitem{lee2018dart}
J.~Lee, M.~X. Grey, S.~Ha, T.~Kunz, S.~Jain, Y.~Ye, S.~S. Srinivasa,
  M.~Stilman, and C.~K. Liu, ``Dart: Dynamic animation and robotics toolkit,''
  \emph{Journal of Open Source Software}, vol.~3, no.~22, p. 500, 2018.

\end{thebibliography}

\end{document}